# Removing Motion Artifact in MRI by Using a Perceptual Loss Driven Deep Learning Framework

Ziheng Guo*[1,2], Danqun Zheng*[3], Shuai Li*[3], Chengwei Chen[3], Boyang Pan[4], Xuezhou Li[3], Ziqin Yu[3], Langdi Zhong[4], Chenwei Shao#[3], Yun Bian#[3], Nan-Jie Gong#[1,2]

[1]Institute of Magnetic Resonance and Molecular Imaging in Medicine, East China Normal University, Shanghai, China

[2]Shanghai Key Laboratory of Magnetic Resonance, School of Physics, East China Normal University, Shanghai, China

[3]The First Affiliated Hospital of Naval Medical University Department of Radiology, Shanghai, China

[4]Laboratory for Intelligent Medical Imaging, Tsinghua Cross-Strait Research Institute, Xiamen, China

[5]RadioDynamic Medical, Shanghai, China

#: these authors are corresponding authors. E-mails: chenweishaoch@163.com, bianyun2012@foxmail.com, nanjie.gong@gmail.com

*: these authors contribute equally

# Abstract

**Purpose:** Deep learning-based MRI artifact correction methods often demonstrate poor generalization to clinical data. This limitation largely stems from the inability of deep learning models in reliably distinguishing motion artifacts from true anatomical structures, due to insufficient awareness of artifact characteristics. To address this challenge, we proposed PERCEPT-Net, a deep learning framework that enhances structure preserving and suppresses artifact through dedicated perceptual supervision.

**Method:** PERCEPT-Net is built on a residual U-Net backbone and incorporates three auxiliary components. The first multi-scale recovery module is designed to preserve both global anatomical context and fine structural details, while the second dual attention mechanisms further improve performance by prioritizing clinically relevant features. At the core of the framework is the third Motion Perceptual Loss (MPL), an artifact-aware perceptual supervision strategy that learns generalized representations of MRI motion artifacts, enabling the model to effectively suppress them while maintaining anatomical fidelity. The model is trained on a hybrid dataset comprising both real and simulated paired volumes, **and** its performance is validated on a prospective test set using a combination of quantitative metrics and qualitative assessments by experienced radiologists.

**Result:** PERCEPT-Net outperformed state-of-the-art methods on clinical data. Ablation studies identified the Motion Perceptual Loss as the primary contributor to this performance, yielding significant improvements in structural consistency and tissue contrast, as reflected by higher SSIM and PSNR values ($p < 0.001$). These findings were further corroborated by radiologist evaluations, which demonstrated significantly higher diagnostic confidence in the corrected volumes (median score: 3 vs. 2, $p < 0.001$).

**Conclusion:** The proposed PERCEPT-Net can effectively suppress motion artifacts in MRI data while preserving anatomical integrity. By overcoming the limitation of over-smoothing and structural degradation, the framework provides a practical solution to robustly remove MRI motion artifact in real clinical settings.

# 1. Introduction

Magnetic resonance imaging (MRI) is an indispensable diagnostic modality in neuroradiology. Owing to its exceptional soft-tissue contrast and high spatial resolution, MRI remains the gold standard for evaluating a wide range of neurological disorders, including brain tumors, Alzheimer's disease, and cerebrovascular pathologies [1–3]. However, its diagnostic performance is frequently degraded by motion artifacts induced by involuntary patient movement during data acquisition. Such artifacts are particularly prevalent in vulnerable populations, including pediatric, geriatric, and neurologically impaired patients, who often struggle to remain still throughout relatively long scan durations [4–7].

Motion artifacts typically manifest as blurring, ghosting, ringing, and reduced signal-to-noise ratio (SNR), which can obscure subtle anatomical structures such as small lesions and fine vascular details, thereby compromising diagnostic reliability [3,4,10,11]. To address this issue, conventional approaches have relied on physical immobilization, accelerated pulse sequences [8], and non-Cartesian k-space sampling strategies [9]. Although non-Cartesian trajectories improve robustness to motion, they generally lack explicit constraints to enforce physical consistency of the acquired signals.

This limitation has motivated the development of physics-driven reconstruction methods [9], which incorporate MRI acquisition models and variational regularization to improve data fidelity. While these methods can produce anatomically plausible reconstructions, they are often computationally expensive and exhibit limited flexibility in handling the complex, non-rigid motion patterns commonly observed in clinical practice.

In parallel, data-driven approaches have emerged to bridge the gap between simulated and real-world clinical scenarios. In particular, unpaired learning frameworks that utilize motion-corrupted and artifact-free scans have been investigated to eliminate the need for strictly aligned data. However, such methods often suffer from distribution mismatch, mode collapse, and ambiguous structural correspondences [23,30].

More recently, generative models, especially diffusion models, have demonstrated strong performance in high-fidelity image restoration tasks [12,13,25,30], benefiting from their iterative denoising process. Despite their effectiveness, these models typically require extensive training time, may hallucinate anatomical structures under severe corruption, and lack explicit mechanisms to disentangle motion artifacts from true tissue signals.

A fundamental challenge shared across these approaches is the difficulty in reliably distinguishing motion artifacts from genuine anatomical structures. This ambiguity often leads to over-smoothing, loss of fine anatomical detail, or even generation of pseudo-structures [15,24]. A key underlying reason is the reliance on generic perceptual supervision derived from natural images, which lacks MRI-specific sensitivity to artifact characteristics and therefore generalizes poorly to clinical data [28,29].

To address this limitation, we propose PERCEPT-Net, a generative deep learning framework centered on a novel Motion Perceptual Loss (MPL). Unlike conventional perceptual losses that capture generic visual similarity, MPL explicitly learns motion-artifact-aware representations, enabling it to enhance the discriminability between artifact-induced distortions and true anatomical structures. By strengthening this separation, the proposed framework suppresses motion artifacts while preserving fine structural details. Furthermore, PERCEPT-Net incorporates multi-scale feature learning and dual-attention mechanisms to further improve anatomical fidelity and robustness in clinical scenarios.

# 2. Method

## 2.1 Data Acquisition and Preparation

### 2.1.1 Multi-Center Data Collection

A prospective cohort of 664 patients was recruited from three medical centers: Changhai Hospital (Shanghai, China), 411 Hospital (Shanghai, China), and Putian Hospital (Fujian, China). 1.5T and 3T MRI scanners with standardized T1-weighted and T2-weighted protocols were used. The study was conducted in accordance with the ethical standards of the institutional review board at each participating center, and comprehensive acquisition parameters are detailed in Table 1.

To capture authentic artifact patterns, the study utilized a hybrid dataset incorporating real pair volumes obtained through rapid re-scanning within 30 minutes of identifying motion artifacts, alongside extensive simulated pair volumes stratified by multiple artifact severity levels. All data partitioning was performed at a strict patient level to prevent information leakage. For stratified analysis, artifact severity was independently categorized by two senior radiologists (≥10 years of experience) into three levels: mild (subtle degradation without obscuring landmarks like the gray-white matter junction), moderate (discernible blurring/ghosting while maintaining diagnostic visibility), and severe (extensive distortion rendering structures like the brainstem uninterpretable), as illustrated in Figure 1.

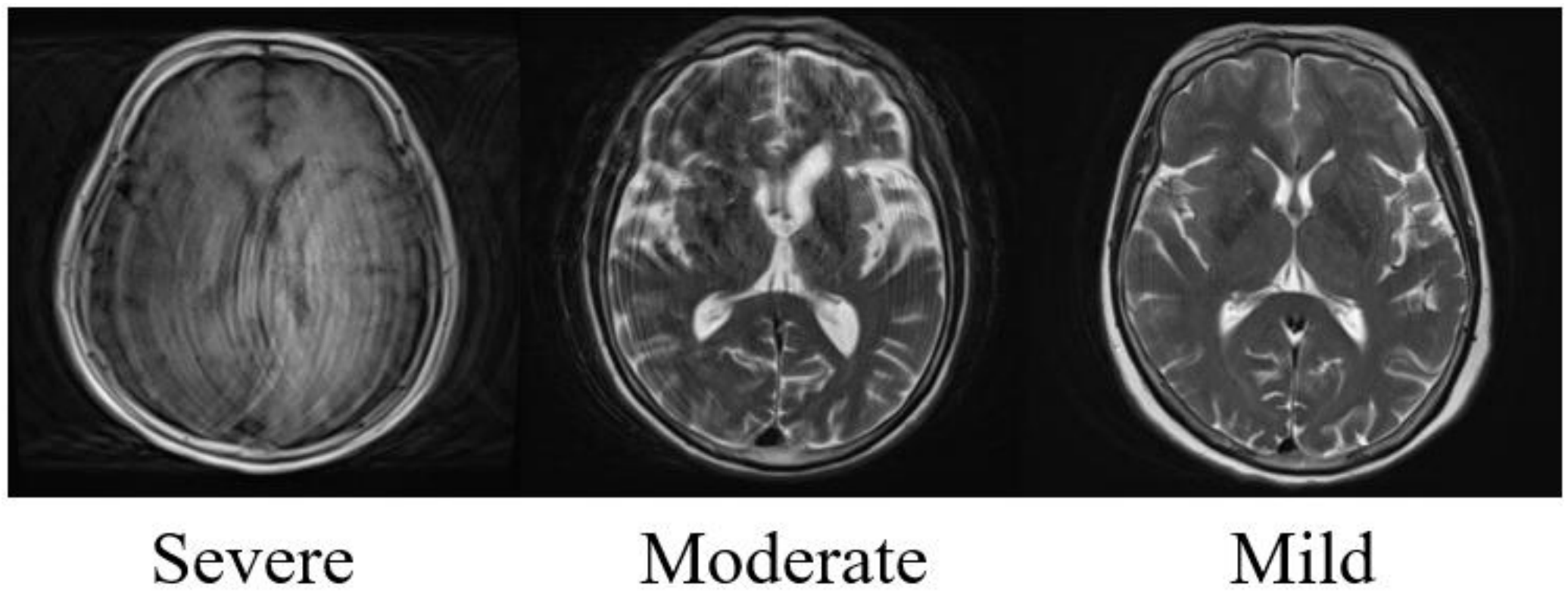

**Figure 1. Representative examples of motion artifact severity. Severe: Extensive distortion and signal loss render key anatomical details. Moderate: Discernible blurring and ghosting are visible, but essential structures remain diagnostically interpretable. Mild: Subtle image degradation is present but does not obscure critical anatomical landmarks or reduce diagnostic confidence.**

### 2.1.2 Multi-Center Dataset Allocation and Preprocessing

The comprehensive multi-center dataset, totaling 12,110 volumes (Figure 2), is organized in a hierarchical structure to facilitate robust artifact representation learning by integrating real-world clinical scans with synthetic samples. The primary cohort from Changhai Hospital comprised 336 real paired volumes and 10,496 unpaired volumes. The real paired volumes were partitioned into Internal Test Set 1 (n=160) and Main Training Set (n=176). Within the unpaired volumes, 784 volumes were used to generate a Simulated Paired Subset, 856 volumes were reserved for Motion Perceptual Loss (MPL) Training, and the remaining 8,856 volumes formed Internal Test Set 2 for large-scale validation. To evaluate cross-center generalizability, two independent external cohorts were established: External Test Set 1, comprising 165 real paired volumes from 411 Hospital, and External Test Set 2, comprising 1,113 unpaired volumes from Putian Hospital. All volumes underwent maximum intensity normalization to standardize the dynamic range across centers and acquisition protocols.

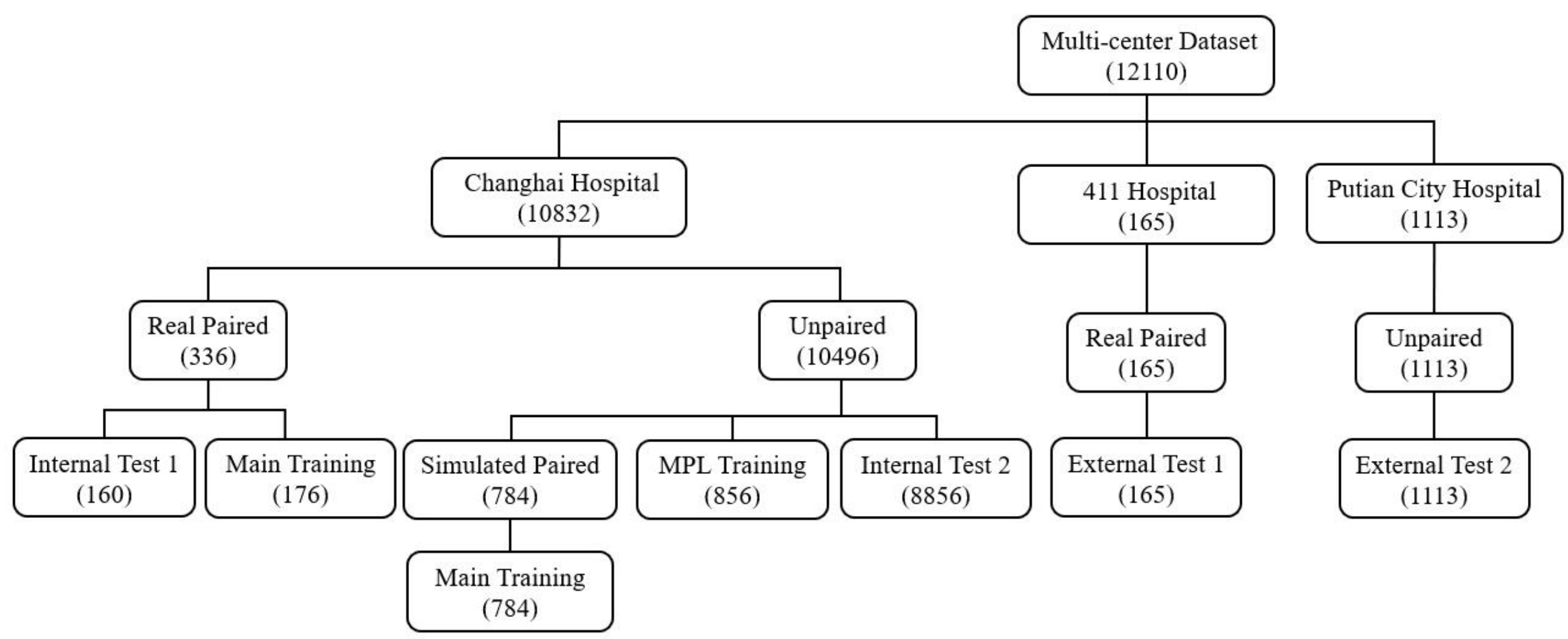

**Figure 2. Overview of the construction of multi-center dataset and training/testing cohorts.**

### 2.1.3 Data Preprocessing for MPL (Motion Perceptual Loss) training

The Motion Perceptual Loss (MPL) training utilized a balanced subset derived from the unpaired data pool, consisting of 428 real motion-corrupted volumes (positive class, label=1) and 428 artifact-free volumes (negative class, label=0). Random undersampling was applied to maintain a strict 1:1 ratio, resulting in a balanced 856-image corpus free from training bias. To preserve the morphological integrity of clinical artifacts, no additional data augmentation was applied to this subset. The cohort was further divided via stratified sampling into a Training Split (70%, 600 images) and a Validation Split (30%, 256 images). Critically, these 856 volumes were entirely isolated from the main training set, serving solely to enable the feature extractor to differentiate authentic motion patterns from normal anatomical structures.

### 2.1.4 Simulated Paired Data Generation

To broaden the network's exposure to varied artifact morphologies, 784 simulated paired volumes were generated from 470 artifact-free volumes using a k-space phase perturbation pipeline [26, 27]. Artifact-free volumes were first transformed into the k-space domain via an Inverse Fast Fourier Transform (IFFT). Rigid motion was then simulated by applying random phase noise at three severity levels: mild ($\pm 0.1\pi$), moderate ($\pm 0.3\pi$), and severe ($\pm 0.5\pi$), combined with uniform undersampling at corresponding ratios (60%–80%, 40%–60%, and 20%–40%). The perturbed data were reconstructed back to the image domain using a Fast Fourier Transform (FFT) and normalized to ensure clinical consistency. As illustrated in Figure 3, these synthetic pairs were integrated into the Main Training Set while remaining strictly isolated from the MPL training set, preserving the specificity of motion artifact features.

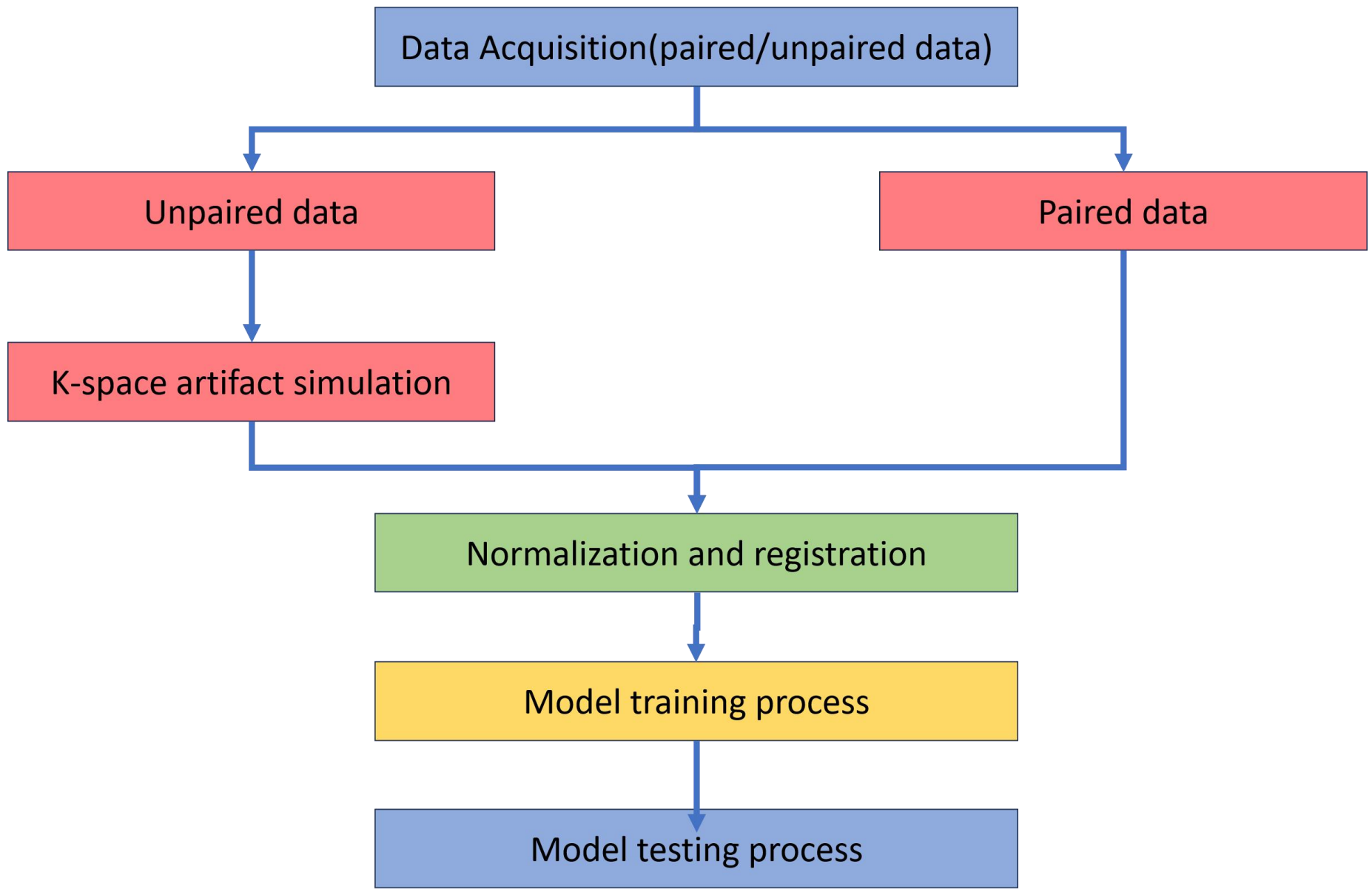

**Figure 3. Overview of the hybrid dataset construction. The main training set comprises real clinical paired volumes and simulated artifact pairs generated by k-space perturbation, with unified intensity normalization, spatial registration, and data augmentation for standardized model training.**

### 2.1.5 Final Dataset Partitioning Strategy

To ensure a rigorous evaluation, all data partitioning followed a strict patient-level, maintaining entirely disjoint subsets to eliminate potential data leakage. The primary Changhai Hospital cohort was categorized into four functional subsets: (1) a Main Training Set of 960 paired volumes (176 augmented real pairs and 784 simulated pairs), split 70:30 for training and validation; (2) an MPL Training Set of 856 unpaired volumes for backbone fine-tuning; and (3) two internal evaluation sets, Internal Test Set 1 (n=160 real pairs) and Internal Test Set 2 (n=8,856 unpaired volumes). Along with the two external test cohorts from 411 Hospital and Putian Hospital, this structured partitioning encompasses diverse institutional sources and unseen patient data to verify the model's clinical robustness (Figure 2).

## 2.2 Network Architecture

The Perceptual Loss-Powered Artifact Removal Network (PERCEPT-Net) is built upon a residual U-Net [22] backbone (Figure 4) with a symmetric encoder-decoder topology, Residual blocks are incorporated into both contracting (encoder) and expanding (decoder) paths to facilitate gradient flow and mitigate the vanishing gradient problem during deep network training. To address the primary challenges of MRI motion artifact correction, three synergistic components are integrated: a multi-scale recovery module, dual attention mechanisms, and the motion perceptual loss (MPL). Below is a detailed elaboration of the architecture design, including structural parameters and implementation details.

### 2.2.1 Multi-scale Recovery Module

A multi-scale recovery (MS) module is embedded in decoding stage to handle anatomical scale variations, such as the differences between large brain lobes and small deep nuclei. This module utilizes parallel convolutions with kernel sizes of 3×3, 5×5, and 7×7 to aggregate features from multiple receptive fields. The output feature maps are concatenated and compressed using a 1×1 convolution for feature fusion. This design enables the simultaneous capture of fine-grained textures including periventricular ependymal lining and global contextual information such as brainstem morphology.

### 2.2.2 Dual Attention Mechanisms

A multi-scale attention (MSA, Figure 5) block integrates channel attention and spatial attention mechanisms to prioritize clinically important regions. The channel attention module adopts a squeeze-and-excitation structure, using global average pooling followed by fully connected layers with ReLU and Sigmoid activations to generate adaptive channel-wise weights. The spatial attention module is applied after the channel attention module and computes spatial attention maps using channel-wise maximum and average pooling, followed by a 3 × 3 convolution and Sigmoid activation. This process highlights salient anatomical regions including deep nuclei and cerebellar peduncles.

### 2.2.3 Motion Perceptual Loss

A central component of our framework is the Motion Perceptual Loss (MPL), specifically designed to encourage the network to learn discriminative feature distributions of authentic motion artifacts for targeted suppression. The MPL utilizes a VGG19 backbone pre-trained on ImageNet, which is subsequently fine-tuned on a specialized dataset consisting exclusively of real motion-corrupted volumes (positive class, label = 1) and artifact-free volumes (negative class, label = 0). This fine-tuning adapts the pre-trained weights to the distinctive morphological characteristics of MRI motion artifacts and underlying anatomical structures.

To optimize this adaptation, we implement a stage-wise freezing strategy that balances the preservation of generic visual descriptors with the acquisition of MRI-specific features. The first ten convolutional layers (spanning the Conv1 to Conv3 blocks) are fully frozen to retain universal low-to-mid-level features [36], such as edges and textures, which are highly transferable and help mitigate overfitting on the limited MRI dataset. Conversely, the high-level convolutional blocks (Conv4 and Conv5, layers 11 through 16) and the subsequent fully connected (FC) layers are unfrozen. These deeper layers are tasked with capturing the complex semantic representations required to distinguish real motion artifacts from simulated patterns and normal anatomy.

Fine-tuning is executed as a binary classification task (discriminating motion-corrupted from artifact-free volumes) using a mini-batch size of 16. The network is optimized via Stochastic Gradient Descent (SGD) with a momentum of 0.9 and a weight decay of $5\times10^{-4}$ . The learning rate is initialized at $1\times10^{-4}$ and decayed by a factor of 0.1 every 10 epochs over a 50-epoch regime, with early stopping (patience of 8 epochs) employed to ensure optimal generalization. Following this process, the fine-tuned VGG19 achieved robust performance on a balanced validation subset (n=256), yielding a validation accuracy of 94.2%, precision of 93.8%, recall of 94.5%, and an F1-score of 94.1%.

Upon completion of the fine-tuning phase, the FC classification head is discarded, and the remaining convolutional backbone is repurposed as a task-specific feature extractor. By shifting the focus from explicit classification to high-level feature representation, the backbone becomes sensitive to clinical motion patterns. The final MPL is computed as the L1 distance between high-dimensional feature maps extracted from the Conv4_4 and Conv5_4 layers, which are selected for their proficiency in balancing mid-to-high level [36] structural details with artifact pattern recognition. The mathematical formulation of the MPL is as follows:

$$\mathcal{L}_{MPL}=\frac{1}{H_l W_l C_l}\sum_{l\in\{4_4,5_4\}}\sum_{i=1}^{H_l}\sum_{j=1}^{W_l}\sum_{k=1}^{C_l}\left|\Phi_l\left(\hat{I}\right)_{i,j,k}-\Phi_l\left(I_{gt}\right)_{i,j,k}\right|$$

where:

- $\phi_l(\cdot)$ denotes the feature extraction function of the l-th target layer (Conv4_4/Conv5_4) of the fine-tuned VGG19;
- $\hat{I}$ is the reconstructed MRI image output by the PERCEPT-Net generator;
- $I_{gt}$ is the artifact-free ground-truth MRI image;

- $H_l W_l C_l$ represent the height, width, and number of channels of the feature map extracted from the l-th layer, respectively;
- The triple summation computes the pixel-wise L1 distance between the feature maps of the reconstructed and ground-truth volumes, and the average over $H_l W_l C_l$ normalizes the loss by the size of the feature map to eliminate scale effects.

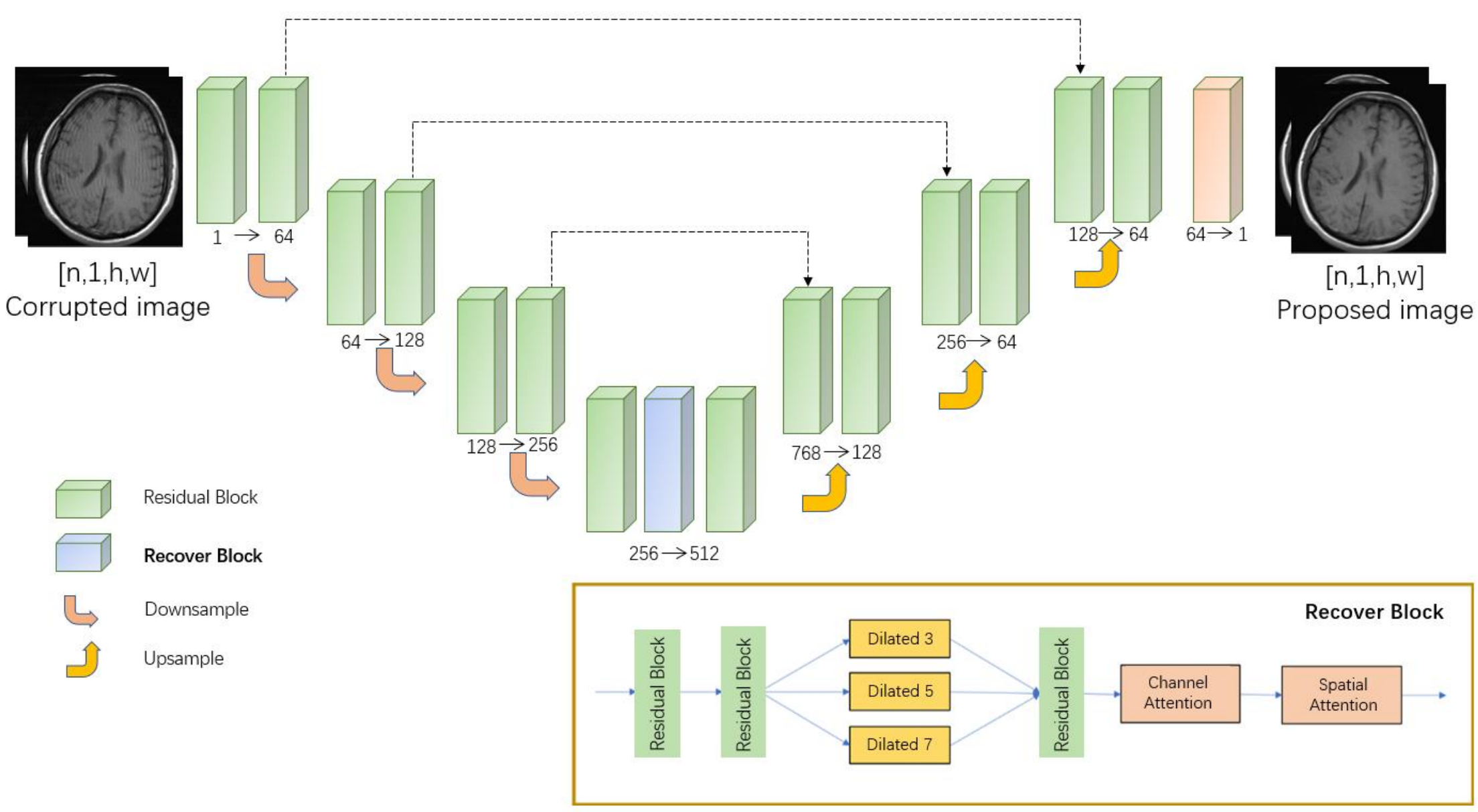

**Figure 4. Illustration of PERCEPT-Net architecture. Structural details were described in paragraph 2.2 Network Architecture.**

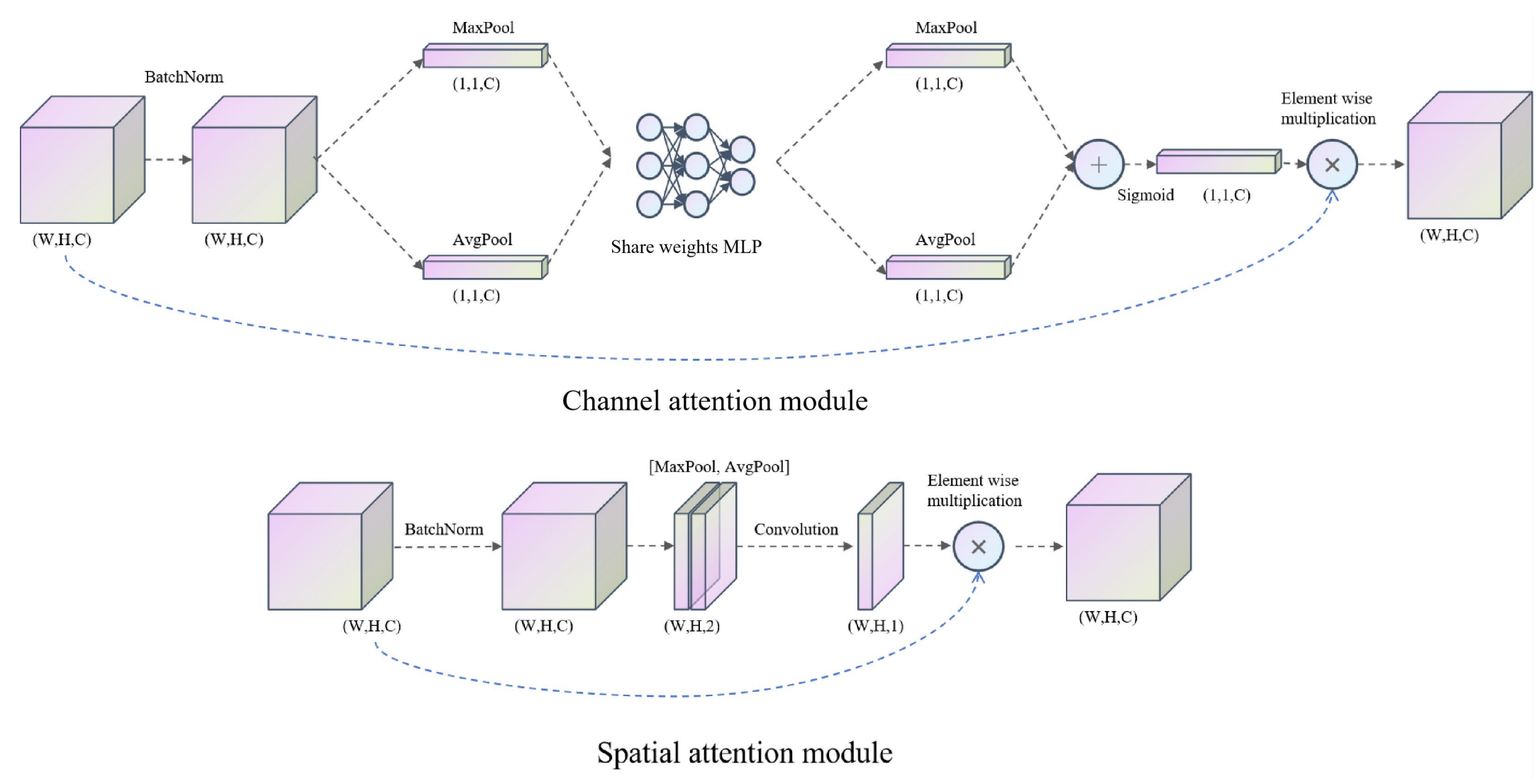

**Figure 5. The overview of channel attention module and spatial attention module.**

## 2.3 Loss Function

The overall objective function is a weighted combination of multiple complementary loss terms, with the MPL as the core component for distinguishing motion artifacts from anatomical structures, thereby effectively suppressing clinical artifacts:

$$\mathcal{L}_{total} = \alpha_{L1} \cdot \mathcal{L}_{L1} + \alpha_{SSIM} \cdot \mathcal{L}_{SSIM} + \alpha_{motion} \cdot \mathcal{L}_{motion} + \alpha_{focal} \cdot \mathcal{L}_{focal} + \alpha_{adv} \cdot \mathcal{L}_{adv}$$

$\mathcal{L}_{L1}$ is the pixel-wise L1 loss that ensures low-level similarity. $\mathcal{L}_{SSIM}$ represents the structural similarity loss to enforce perceptual consistency in structural details. $\mathcal{L}_{motion}$ is the MPL. $\mathcal{L}_{focal}$ denotes the focal frequency loss [31], which operates in the k-space domain to improve reconstruction of frequency components with higher perceptual importance. $\mathcal{L}_{adv}$ is the adversarial loss [32] imposed by the discriminator to guide the generator toward producing realistic outputs. In addition, $\alpha_{L1}, \alpha_{SSIM}, \alpha_{motion}, \alpha_{focal}, \alpha_{adv}$ serve as weight hyperparameters that regulate the contribution of each corresponding loss component to the total loss. All these hyperparameters are constrained to the range (0,1] and their sum is usually normalized to 1 (i.e., $\alpha_{L1} = 0.25, \alpha_{SSIM} = 0.25, \alpha_{motion} = 0.3, \alpha_{focal} = 0.15, \alpha_{adv} = 0.05$) to ensure the rational distribution of optimization focus across different loss components.

PERCEPT-Net was implemented in PyTorch (version 2.0.1) and trained on an NVIDIA RTX 3090 with 24 GB GPU RAM. The model was trained for 100 epochs with a batch size of 64, using the Adam optimizer with an initial learning rate of $1\times10^{-4}$ . A cosine annealing scheduler was employed to adjust the learning rate dynamically, with a minimum learning rate to $1\times10^{-5}$ . The total training time was approximately two days. For inference, the model was deployed on an NVIDIA RTX A4000 with 16 GB GPU RAM, achieving an average inference time of 4 seconds per volume (size: 30×512×512).

## 2.4 Image Quality Assessment

### 2.4.1 Quantitative Evaluation and Statistical analysis

This prospective research utilized a mixed dataset derived from a multi-center cohort. To evaluate the performance of the model, several quantitative metrics were employed: PSNR and SSIM were used to assess pixel-level accuracy and structural similarity, while SNR and CNR measured signal and contrast stability. Additionally, LPIPS [34] and FID [35] were calculated to evaluate deep perceptual similarity and distribution matching between the corrected volumes and the ground truth.

Regarding statistical analysis, the normality and variance homogeneity of the data were first verified. Subsequently, a paired t-test was performed for normally distributed data, whereas the Wilcoxon signed-rank test was applied for data exhibiting non-normal distributions or unequal variance. To control for Type I errors arising from multiple comparisons, False Discovery Rate (FDR) correction was implemented within each family of tests, defined as all model comparisons for a specific metric within a given test dataset. Statistical significance is denoted as follows: *p < 0.05, **p < 0.01, ***p < 0.001, and ns for non-significant results. All quantitative findings are summarized in Table 4 for ablation studies and Table 5 for artifact severity stratification.

#### 2.4.2 Qualitative Radiologist Evaluation

Subjective image quality was independently assessed by two experienced radiologists, each with over ten years of expertise in neuro-MRI interpretation, who remained blinded to the experimental groups. Each MRI volume was evaluated using a 5-point Likert scale, where scores ranged from 1 (unacceptable) to 5 (excellent). As detailed in Table 2, seven key anatomical structures were specifically targeted: the gray-white matter junction (A1), deep nuclei (A2), brainstem (A3), internal auditory canal (A4), suprasellar cistern (A5), periventricular region (A6), and cerebellum (A7). Inter-rater reliability between the two readers was quantified using the intraclass correlation coefficient (ICC). All qualitative results are reported as median (interquartile range, IQR) in Table 6.

# 3. Results

## 3.1 Ablation study

Three complementary ablation studies were conducted to systematically validate the efficacy of the proposed framework: first, to assess the impact of training data composition by isolating simulated versus real clinical data sources, thereby elucidating the value of authentic artifact representations; second, to verify the specific contribution of the MPL in simultaneously preserving anatomical integrity and suppressing motion-induced artifacts; and third, to evaluate the synergistic effect of integrating multi-scale recovery, dual attention mechanisms, and the Motion Perceptual Loss within the unified architecture.

#### 3.1.1 Ablation Study of Training Data Composition

To assess the impact of different training data types on artifact correction performance and provide a baseline for evaluating the MPL, we trained two baseline ResUNet variants: ResUNet-Sim, which was trained exclusively on simulated motion artifact data, and ResUNet-Real, which was trained exclusively on real clinical motion artifact data to capture the complex patterns of authentic patient motion and provide a clinically relevant performance benchmark. Table 4 & 5 present quantitative results comparing the two baseline models which isolate the effect of data authenticity and directly address the core challenge of distinguishing real from simulated artifacts.

On datasets with ground-truth (Internal Test 1: Changhai Hospital; External Test 1: 411 Hospital), ResUNet-Real achieved consistently more favorable results than ResUNet-Sim across key quantitative metrics. In the Changhai Hospital with GT cohort, ResUNet-Real attained a higher SSIM (and 0.589 vs. 0.569), a higher CNR (40.170 vs. 35.351), a higher SNR (25.573 vs. 24.704), and a lower FID (7.431 vs. 11.162) and lower LPIPS (0.217 vs. 0.275). Similarly, in the 411 Hospital cohort, ResUNet-Real achieved a higher SSIM (0.642 vs. 0.624) and lower LPIPS (0.195 vs. 0.245) compared with ResUNet-Sim. These results confirm that training on real clinical data enables the model to capture complex artifact patterns that simulated data cannot adequately reproduce.

On datasets without ground-truth (Changhai Hospital without GT and Putian City Hospital), where SNR and CNR serve as quality surrogates, ResUNet-Real generally maintained superiority. It showed higher SNR and CNR, despite a slight SNR trade-off in the latter attributable to simulated data's-controlled noise. Collectively, training with real clinical data enhances image contrast and perceptual realism.

### 3.1.2 Ablation Study of Motion Perceptual Loss Function

To validate the core innovation of the proposed framework, a targeted ablation study was conducted, focusing exclusively on the impact of the MPL. In contrast to Section 3.1.1, where models were trained separately on simulated or real data, all models in this experiment were trained on the hybrid training set consisting of both real clinical paired volumes and simulated paired volumes, which is fully consistent with the main training set defined in Section 2.1.5. Using the baseline ResUNet [22] as the foundational architecture, two model variants were compared to directly assess the specific contribution of the MPL, independent of other architectural modifications:

- ResUNet without MPL: The baseline ResUNet model trained without incorporating the MPL.
- ResUNet with MPL: The baseline ResUNet model enhanced by integrating the proposed MPL.

The quantitative results, summarized in Table 4, directly demonstrate the efficacy of the MPL in suppressing authentic motion artifacts while preserving anatomical integrity. On datasets possessing ground-truth references, the integration of the MPL yielded statistically significant enhancements across most primary quantitative metrics.

For the 411 Hospital cohort (External Test 1), ResUNet augmented with the MPL demonstrated a 3.8% relative improvement in SSIM, achieving a value of 0.680(vs 0.663 for ResUNet without MPL, relative improvement = 2.6%), an SNR of 1.049 (vs. 1.040), and an LPIPS of 0.197(vs. 0.205), compared to the baseline model without MPL. For the Changhai Hospital with GT cohort (Internal Test 1), ResUNet with MPL yielded an SSIM of 0.664 (vs. 0.613, relative improvement = 8.3%), a CNR of 40.484 (vs. 34.094), and an FID of 7.345 (vs. 7.543). As shown in Figure 6, ResUNet with MPL demonstrates superior performance in motion artifact removal while better preserving structural information. The consistent performance gains across independent clinical datasets validate the MPL's efficacy in enabling the network to discriminate the complex patterns of genuine motion artifacts from underlying anatomical structures.

In the absence of paired ground-truth references, SNR and CNR served as surrogate indicators for image clarity and tissue contrast preservation. The beneficial impact of the MPL remained evident under these conditions. For Putian City Hospital dataset (External Test 2), ResUNet with MPL achieved an SNR and a CNR values of 33.381 and 57.530. In the Changhai Hospital without GT dataset (Internal Test 2), ResUNet with MPL achieved an SNR of 26.549 (+2.4%) and a CNR of 35.054 (+4.8%). These results underscore the effectiveness of the MPL in enhancing perceived image quality even when ground truth is unavailable for direct optimization.

Overall, this ablation study confirms that the MPL is the primary driver for the model's improved performance. By explicitly learning the discriminative features of real-world motion artifacts, the MPL enables the network to achieve targeted artifact suppression while maintaining critical anatomical details, thereby addressing a fundamental limitation in generalizing artifact correction models to clinical data.

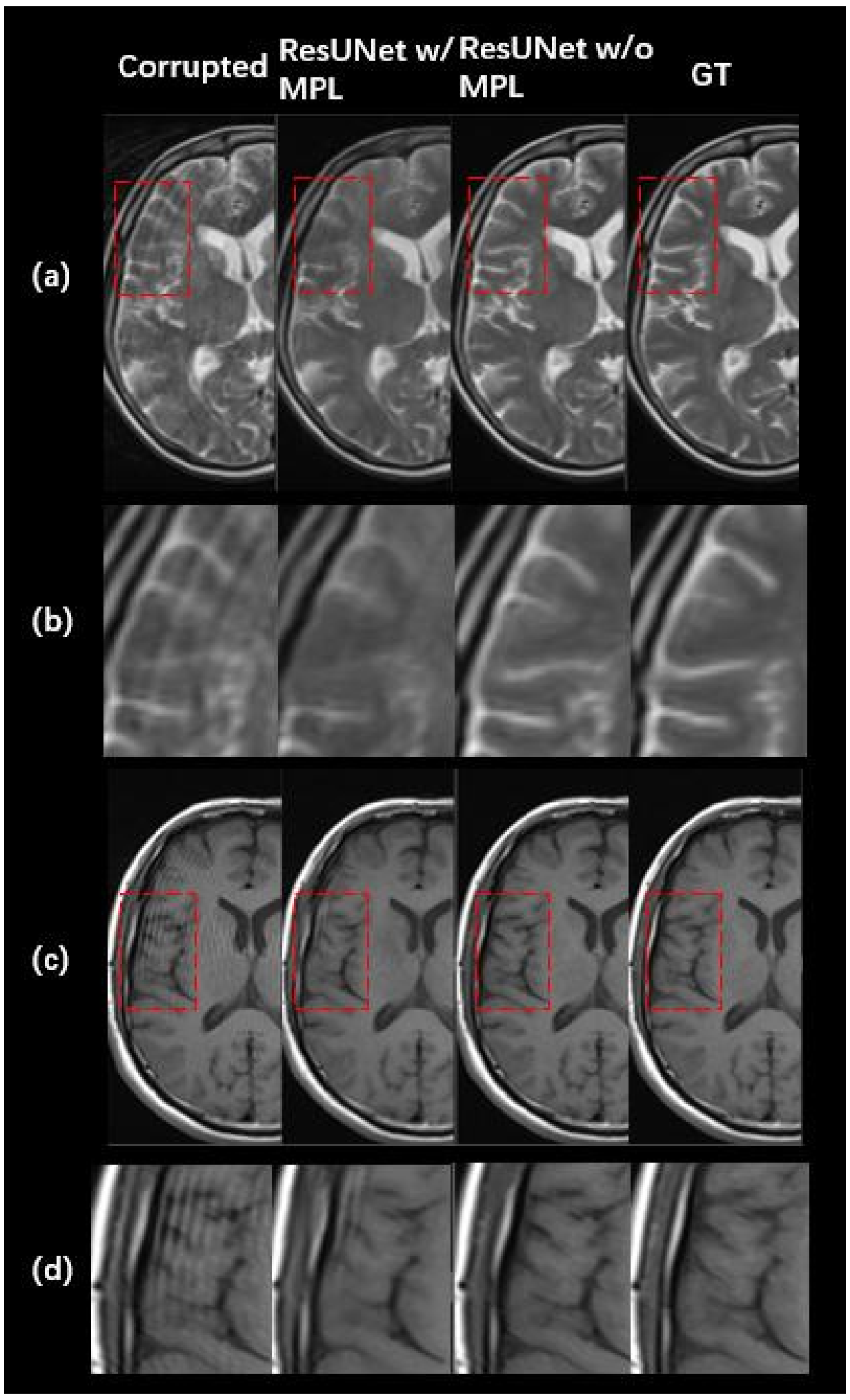

**Figure 6. Ablation study: Visual evidence of motion perceptual loss contribution. Panels (a) and (c) illustrate the effect of MPL on the overall restoration quality, respectively, while panels (b) and (d) show zoomed-in details of the key regions marked by the red boxes for better visualization.**

### 3.1.3 Ablation Study of Efficacy of Integrated Architectural Innovations

This experiment evaluates the synergistic effect of all proposed components. Starting with the baseline ResUNet [22], we incrementally integrated modules and compared performance at each step:

- **MS-ResUNet:** ResUNet extended with a multi-scale recovery module.

- **MSA-ResUNet:** MS-ResUNet further enhanced by integrating dual attention mechanisms (channel and spatial).
- **PERCEPT-Net:** The final MSA-ResUNet architecture, augmented with the proposed MPL.

We also included the state-of-the-art ResShift [24] as a reference. Results in Table 4 demonstrate that PERCEPT-Net achieves competitive and favorable performance across diverse datasets, with consistent performance gains that highlight the synergistic value of integrated components.

On datasets with ground truth, PERCEPT-Net exhibited consistent advantages over the three ablated variants. In the 411 Hospital cohort, PERCEPT-Net achieved an SSIM of 0.678 and CNR of 15.532, which were notably higher than 0.663 and 12.359 for the baseline ResUNet, and also exceeded MS-ResUNet and MSA-ResUNet. Similarly, in the Changhai Hospital with GT cohort, PERCEPT-Net attained the highest SSIM (0.665) and CNR (41.69) among all incremental models, along with the lowest LPIPS (0.202) and FID (7.021). These results confirm that the full framework yields stronger structural consistency and tissue contrast than its simplified counterparts.

On datasets without ground truth, PERCEPT-Net remained stable and competitive. In the Putian City Hospital cohort, it achieved a CNR of 61.06, considerably outperforming ResUNet (52.12), MS-ResUNet (52.58), and MSA-ResUNet (52.92). In the Changhai Hospital without GT cohort, PERCEPT-Net also maintained favorable CNR and SNR, demonstrating reliable generalization to real clinical scans.

The progressive performance improvement from ResUNet to MS-ResUNet, MSA-ResUNet, and finally PERCEPT-Net reveals the complementary roles of each component. The multi-scale module in MS-ResUNet improved fine-grained anatomical representation compared with the baseline. The dual attention mechanism in MSA-ResUNet further strengthened the focus on clinically salient regions. By incorporating the MPL, PERCEPT-Net delivered the most substantial performance gains, especially in CNR and perceptual fidelity, confirming that the MPL serves as the core innovation that enables the model to accurately distinguish real motion artifacts from genuine anatomical structures.

Beyond quantitative metrics, visual comparisons in Figure 7 show that competing models often retain residual artifacts or pseudo-structures in motion-affected regions, especially along the gray–white matter junction and periventricular areas. In contrast, PERCEPT-Net provides sharper structural delineation and more homogeneous tissue contrast while preserving anatomical boundaries. The performance improvements of PERCEPT-Net over all ablated models and ResShift were statistically significant across most key metrics (***$p < 0.001$ for SSIM, CNR, FID, LPIPS; **$p < 0.01$ for PSNR).

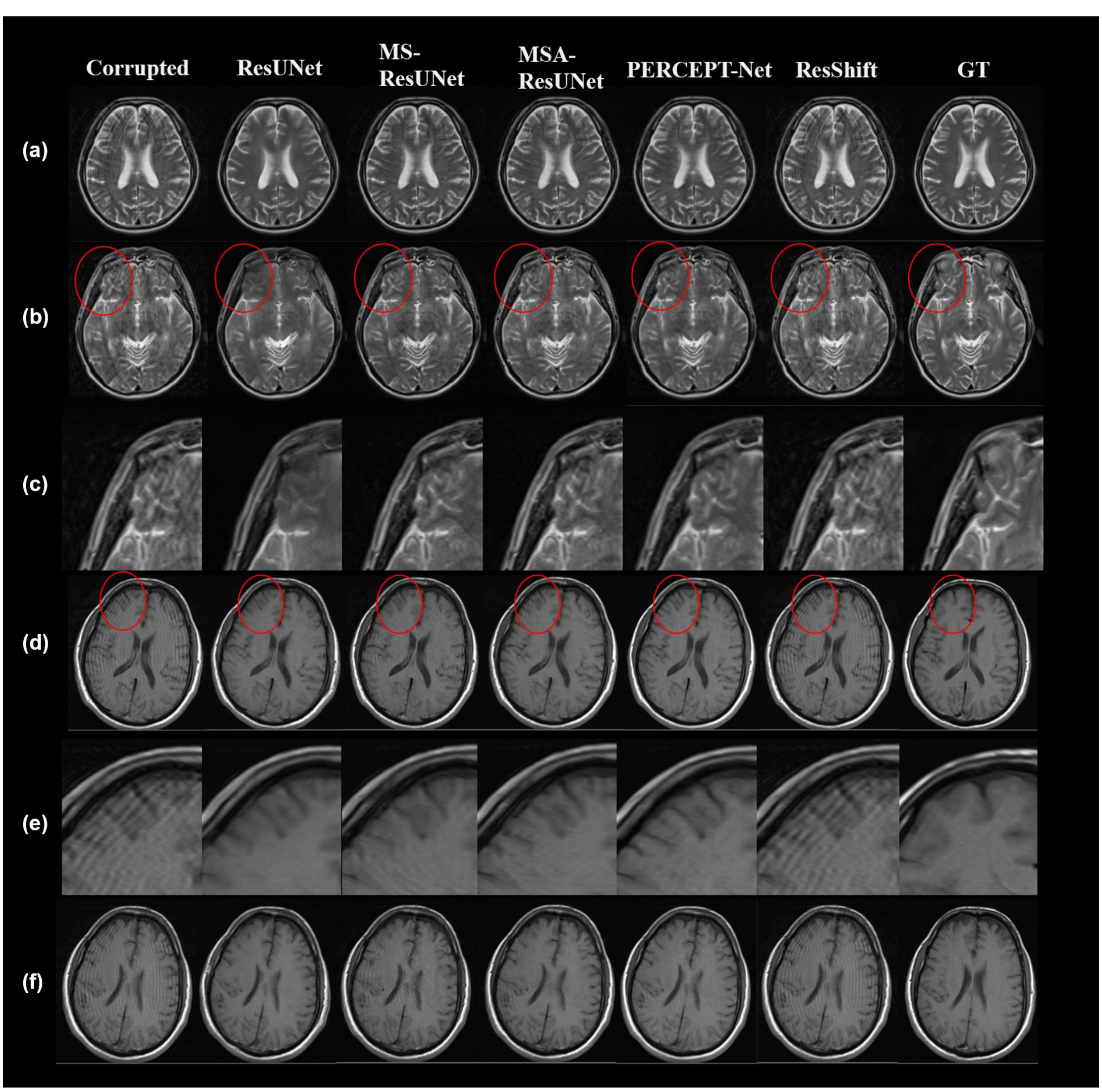

**Figure 7. Visual evidence of integrated architectural innovations. Panels (a) (b) (d) (f) illustrate the motion artifact removal performance under real-world scanning conditions, while panels (c) and (e) show zoomed-in details of the key regions marked by the red circles for better visualization.**

## 3.2 Performance Across Varying Motion Artifact Severities

To evaluate the clinical robustness of PERCEPT-Net, we assessed its performance on volumes stratified by the severity of motion artifacts (mild, moderate, and severe). Experiments were conducted on the Changhai Hospital with GT dataset (Internal Test 1, Table 5), which included 160 volumes (85 T1-weighted, 75 T2-weighted). For T1-weighted volumes, 57 were mild, 14 moderate, and 14 severe. For T2-weighted volumes, 11 were mild, 37 moderate, and 27 severe. Overall, PERCEPT-Net exhibited consistent performance improvements that varied logically with artifact severity and sequence type.

For T1-weighted volumes, PERCEPT-Net achieved significant improvements across most metrics at all severity levels. In moderate and severe artifacts, PERCEPT-Net significantly increased SSIM and PSNR (both ***$p < 0.001$), along with significant improvements in LPIPS and FID (***$p < 0.001$). Specifically, SSIM increased from 0.609 to 0.754 in moderate artifacts and from 0.598 to 0.753 in severe artifacts, while CNR improved from 32.261 to 48.026 and from 26.710 to 32.733, respectively. Even in mild T1 artifacts, PERCEPT-Net showed significant gains in SSIM, PSNR (** $p < 0.01$ and *** $p < 0.001$), although absolute improvement was smaller due to the high baseline quality of mildly corrupted volumes.

For T2-weighted volumes, PERCEPT-Net achieved substantial improvements in moderate and severe artifacts, with significant enhancements across most metrics (***$p < 0.001$). SSIM increased from 0.476 to 0.602 in moderate artifacts and from 0.393 to 0.555 in severe artifacts, while CNR improved from 25.558 to 31.633 and from 26.186 to 33.309, respectively. For mild T2 artifacts, PERCEPT-Net still showed significant gains in several key metrics including SSIM, PSNR, CNR, and FID (** $p < 0.01$), as well as in CNR and FID (* $p < 0.05$). This consistent performance, even in high-quality baseline images with subtle motion effects, underscores the model's sensitivity and its robust capacity to further refine perceptual details and signal integrity.

Overall, PERCEPT-Net performs most effectively in moderate-to-severe artifacts, where multi-scale attention and the MPL jointly restore anatomical structure and perceptual fidelity. The MPL enables the network to distinguish genuine motion artifacts from anatomical tissues across both T1 and T2 volumes, supporting stable performance in clinical scenarios. Importantly, in mild artifacts—especially T2-weighted volumes—PERCEPT-Net preserves image quality without generating secondary artifacts, confirming its safety and stability in routine clinical use. The differing magnitudes of improvement between T1 and T2 volumes can be explained by their intrinsic contrast mechanisms and distinct motion artifact characteristics.

### 3.3 Clinical qualitative analysis

Clinical qualitative evaluation of image quality was conducted using a 5-point Likert scale, focusing on the visibility of anatomical structures and overall clarity across seven predefined brain regions (A1–A7). The primary objective was to assess whether the model effectively preserved normal anatomical structures while suppressing genuine motion artifacts. Given the ordinal nature of the Likert scale, statistical analyses were conducted using nonparametric methods, consistent with guidelines for ordinal outcome data.

As shown in Table 6, both the proposed image group and the original standard scanned image group (i.e., Ground Truth, GT) achieved significantly higher scores than the corrupted image group across all regions (all ***p < 0.001). Inter-rater reliability was quantified using the Intra-class correlation coefficient (ICC) with a two-way mixed-effects model and absolute agreement definition, with 95% confidence intervals (CI) reported.

The proposed image group showed good inter-rater reliability, with an ICC of 0.800 (95% CI: 0.750 to 0.840), indicating strong consistency between the two radiologists and supporting the robustness of the subjective assessment.

For the standard scanned image group , the ICC was 0.073 (95% confidence interval: -0.03 to 0.18). This low estimated ICC was statistically expected due to the severely restricted score distribution: both raters assigned uniformly high scores (4 or 5) across nearly all GT volumes, resulting in minimal between-subject variance. Because ICC quantifies the proportion of between-subject variability relative to total variance, such a restricted range inherently reduces the estimated ICC value, rather than reflecting poor actual agreement.

For the corrupted image group, the ICC was 0.417 (95% CI: 0.310 to 0.520). Although both raters generally assigned low scores (< 3), the wider spread of ratings and greater measurement variability in the corrupted volumes contributed to a lower ICC estimate. Notably, ICC estimates must be interpreted with caution when score distributions are highly restricted or skewed, as these conditions can lead to statistical underestimation of true inter-rater agreement rather than indicating poor practical concordance between evaluators.

Across all evaluated regions (A1–A7), the proposed group consistently outperformed the corrupted group, demonstrating uniform enhancement in perceived image quality. Notably, the most substantial improvements were observed in anatomically complex regions such as the gray-white matter junction (A1) and the deep nuclei (A2). This indicates that the MPL effectively identifies and suppresses artifacts in these detail-critical areas, thereby preserving fine anatomical structures and optimizing contrast.

Using a predefined clinical criterion (Likert score ≥ 3 as diagnostically acceptable), 51.45% of the original corrupted volumes fell below the acceptable quality level and would require repeat scanning. After applying our method, the re-scan rate decreased to 27.80%, representing an absolute reduction of 23.65%. These findings demonstrate that the proposed method substantially lowers the clinical re-scan rate, thereby improving workflow efficiency and reducing patient burden.

## 4. Discussion

The generalization gap between synthetic training environments and clinical reality remains a primary hurdle for deep learning-based MRI restoration. This gap often manifests as "artifact-tissue confusion", where models fail to distinguish non-linear motion patterns from authentic anatomical structures. While general-purpose image restoration frameworks like ResShift[24] have demonstrated remarkable generative performance in various inverse problems, their application to clinical MRI still remains quite challenging. Our comparative analysis indicates that although diffusion-based methods can produce visually sharp textures, they may occasionally prioritize stochastic detail generation over the strict preservation of anatomical fidelity. In contrast, PERCEPT-Net is specifically tailored to address these medical imaging nuances. By leveraging deterministic structural constraints rather than purely generative priors, our approach ensures that reconstructed features remain anchored to the underlying anatomy, which is a critical requirement for diagnostic reliability.

The ablation studies (Table 4) provide a granular perspective on the synergy between our proposed modules. The limited performance of the baseline model (ResUNet-Sim) highlights that standard supervised learning on simulated data is insufficient to bridge the domain gap to complex clinical motion. While the Multi-Scale Recovery and Dual Attention modules provided incremental structural refinements, the integration of Motion Perceptual Loss (MPL) was the primary driver of the significant elevation in CNR (from 34.094 to 40.484 in Internal Test 2 set). This suggests that for motion correction, perceiving the specific geometric and perceptual characteristics of artifacts is more vital than simply increasing the depth or complexity of a generic encoder-decoder backbone.

Furthermore, the evaluation across multiple clinical sites (Table 4) underscores the framework's robustness against hardware and protocol variability. Although absolute metrics such as SSIM and PSNR exhibited minor fluctuations between different centers, likely reflecting variations in scanner manufacturers (e.g., UIH vs. Siemens) and sequence configurations, the relative performance gain of PERCEPT-Net remained remarkably consistent. This cross-site stability demonstrates that the perceptual features captured by our model are not overfitted to a specific acquisition environment, but rather represent a generalized understanding of motion-induced signal perturbations.

The clinical utility of our approach is further validated by the performance analysis across artifact severities (Table 5) and radiological scoring (Table 6). PERCEPT-Net exhibited its strongest restorative power in moderate-to-severe cases, where it significantly improved image interpretability. Notably, in the "Mild" category, the model provided statistically significant improvements (** $p < 0.01$) without introducing secondary artifacts or "hallucinations." From a workflow perspective, the observed reduction in the potential re-scan rate from 51.45% to 27.80% demonstrates a tangible path toward optimizing clinical throughput and patient comfort.

Despite its robust performance across T1 and T2 sequences, the differential recovery rates suggest that sequence-specific motion sensitivity warrants further exploration. We acknowledge that extreme bulk motion still presents physical limits to signal recovery; the remaining 27.80% of cases that may still require re-scanning indicate that even with advanced perceptual supervision, some k-space information loss is irreversible. Future iterations of PERCEPT-Net could benefit from incorporating k-space consistency layers to further anchor image-space predictions. Additionally, larger-scale studies focusing on the preservation of subtle pathological lesions are necessary to further establish the clinical safety profile of this approach.

## 5. Conclusion

In this study, we proposed the PERCEPT-Net, a deep learning framework for MRI motion artifact removal. By introducing the Motion Perceptual Loss (MPL), the model learns generalizable representations of real motion artifacts, enabling accurate artifact suppression while preserving anatomical integrity. Validated on multi-center clinical data, PERCEPT-Net achieves superior performance over existing methods and improves diagnostic image quality. This work provides a robust and practical solution for motion artifact removal, promoting the reliability of deep learning-based MRI enhancement and reconstruction in clinical practice.

**Table 1. Detailed acquisition parameters of each sequence across the three centers**

| Center | ChangHai Hospital | | 411 Hospital | | Putian City Hospital | |
|---|---|---|---|---|---|---|
| Modality | T1WI | T1WI | T1WI | T2WI | T1WI | T2WI |
| Device | United-Imaging UMR560 | | Siemens Magnetom Essenza | | Siemens Magnetom Vida | |
| Slice Plane | Axial | Axial | Axial | Axial | Axial | Axial |
| TR(ms) | 4.1 | 4000 | 3.5 | 3000~5000 | 3.8 | 2000~4000 |
| TE(ms) | 2.3 | 85 | 1.2 | 80 | 1.3 | 90 |
| FOV(mm) | 400*400 | 400*400 | 380*380 | 380*380 | 380*380 | 380*380 |
| Matrix | 256*192 | 256*192 | 320*240 | 320*240 | 320*240 | 320*256 |
| Slice Thickness(mm) | 5 | 5 | 3 | 5 | 3 | 5 |
| Slice Gap(mm) | 0 | 0 | 0 | 1 | 0 | 1 |

**Table 2. Radiologist qualitative assessment of image features**

| Number | Region |
|---|---|
| A1 | Gray white matter junction |
| A2 | Deep nuclei (putamen, globus pallidus, internal capsule) |
| A3 | Brainstem (midbrain, pons, fourth ventricle) |
| A4 | Internal auditory canal (facial cochlear nerve complex, CSF) |
| A5 | Suprasellar cistern (optic chiasm, Willis circle) |
| A6 | Periventricular region (ependymal lining, CSF signal) |
| A7 | Cerebellum (hemisphere, vermis, peduncles) |
| Global | Global image confidence |

**Table 3. 5-point Likert scale for image quality, lesion conspicuity, and image sharpness**

| Score | Lesion Conspicuity | Diagnostic confidence |
|---|---|---|
| 1 | Invisible and extremely difficult to identify | Poor diagnostic confidence |
| 2 | Difficult to identify but some details are recognizable | Limited diagnostic confidence |
| 3 | Moderate clarity, with most details identifiable | Moderate diagnostic confidence |
| 4 | Clearly visible, with details easily recognizable | Good diagnostic confidence |
| 5 | Extremely clear, with all details very easily identifiable | Excellent diagnostic confidence |

**Table 4. Averaged results of the Ablation Study (411 Hospital Changhai Hospital with GT, Putian City Hospital and Changhai Hospital without GT). ns: not significant, *:p < 0.05, **:p < 0.01, ***:p < 0.001. ResUNet* denotes ResUNet w/o MPL. For datasets with ground truth, metrics (PSNR, SSIM, CNR, SNR, LPIPS, FID) were computed between model outputs and ground truth. For datasets without ground truth, only SNR and CNR were calculated directly on the volumes. All significance markers indicate comparisons between each model and the corrupted input.**

| | | 411 Hospital | Changhai Hospital with GT | Putian City Hospital | Changhai Hospital without GT |
|---|---|---|---|---|---|
| PSNR | GT | +∞ | +∞ | | |
| | Corrupted | 20.968 | 18.712 | | |
| | PERCEPT-Net | 21.721 (**) | 20.731 (***) | | |
| | MSA-ResUNet | 20.883 (*) | 20.694 (**) | | |
| | MS-ResUNet | 20.907 (*) | 18.756 (*) | | |
| | ResUNet* | 20.424 (***) | 18.624 (**) | | |
| | ResShift | 21.438 (***) | 18.713 (***) | | |
| | ResUNet-Sim | 21.374 (***) | 18.606 (***) | | |
| | ResUNet-Real | 20.864 (***) | 19.772 (ns) | | |
| | ResUNet w/ MPL | 21.184 (**) | 18.722 (**) | | |
| SSIM | GT | 1.0 | 1.0 | | |
| | Corrupted | 0.620 | 0.562 | | |
| | PERCEPT-Net | 0.678 (***) | 0.665 (***) | | |
| | MSA-ResUNet | 0.644 (***) | 0.655 (***) | | |
| | MS-ResUNet | 0.649 (***) | 0.592 (***) | | |
| | ResUNet* | 0.663 (***) | 0.613 (***) | | |
| | ResShift | 0.622 (***) | 0.558 (***) | | |
| | ResUNet-Sim | 0.624 (ns) | 0.569 (***) | | |
| | ResUNet-Real | 0.642 (***) | 0.589 (***) | | |
| | ResUNet w/ MPL | 0.680 (***) | 0.664 (***) | | |
| CNR | GT | 13.581 | 37.709 | | |
| | Corrupted | 11.621 | 30.429 | 52.020 | 33.039 |
| | PERCEPT-Net | 15.532 (***) | 41.688 (***) | 61.062(***) | 35.063(***) |
| | MSA-ResUNet | 12.167 (ns) | 32.271 (ns) | 52.921(**) | 37.246(***) |
| | MS-ResUNet | 12.323 (***) | 34.932 (*) | 52.578(***) | 37.132(**) |
| | ResUNet* | 12.359 (***) | 34.094 (***) | 52.122(***) | 33.424(***) |
| | ResShift | 11.633 (**) | 30.585 (***) | 52.537(***) | 32.952(***) |
| | ResUNet-Sim | 11.764 (*) | 35.351 (***) | 53.680(**) | 37.335(***) |
| | ResUNet-Real | 12.394 (***) | 40.170 (***) | 54.537(***) | 38.701(***) |
| | ResUNet w/ MPL | 14.955 (***) | 40.484 (***) | 57.530(***) | 35.054(***) |
| SNR | GT | 1.132 | 1.0 | | |
| | Corrupted | 1.070 | 24.400 | 32.408 | 25.699 |
| | PERCEPT-Net | 1.059 (***) | 28.872 (***) | 33.694(***) | 26.320(***) |
| | MSA-ResUNet | 1.065 (***) | 28.438 (***) | 32.007(***) | 26.668(***) |
| | MS-ResUNet | 1.070 (ns) | 25.703 (***) | 32.511(**) | 26.189(***) |
| | ResUNet* | 1.040 (***) | 26.615 (***) | 32.528(***) | 25.921(***) |
| | ResShift | 1.117 (***) | 24.227 (***) | 32.530(***) | 25.862(***) |
| | ResUNet-Sim | 1.033 (***) | 24.704 (***) | 32.533(***) | 27.941(***) |
| | ResUNet-Real | 1.040 (***) | 25.573 (***) | 32.945(***) | 26.211(***) |
| | ResUNet w/ MPL | 1.049 (ns) | 28.828 (***) | 33.381(*) | 26.549(***) |
| FID | GT | 0 | 0 | | |
| | Corrupted | 20.691 | 10.696 | | |
| | PERCEPT-Net | 18.638 (***) | 7.021 (***) | | |
| | MSA-ResUNet | 18.868 (***) | 8.606 (***) | | |
| | MS-ResUNet | 18.500 (***) | 7.376 (***) | | |
| | ResUNet* | 19.203 (***) | 7.543 (***) | | |
| | ResShift | 20.515 (**) | 10.697 (ns) | | |
| | ResUNet-Sim | 19.564 (***) | 11.162 (**) | | |
| | ResUNet-Real | 18.923 (***) | 7.431 (***) | | |
| | ResUNet w/ MPL | 18.543 (***) | 7.345 (***) | | |
| LPIPS | GT | 0 | 0 | | |
| | Corrupted | 0.233 | 0.266 | | |
| | PERCEPT-Net | 0.204 (***) | 0.202 (***) | | |
| | MSA-ResUNet | 0.221 (***) | 0.257 (**) | | |
| | MS-ResUNet | 0.225 (***) | 0.262 (ns) | | |
| | ResUNet* | 0.205 (***) | 0.236 (***) | | |
| | ResShift | 0.237 (***) | 0.266 (*) | | |
| | ResUNet-Sim | 0.245 (***) | 0.275 (***) | | |
| | ResUNet-Real | 0.195 (***) | 0.217 (***) | | |
| | ResUNet w/ MPL | 0.197 (***) | 0.211 (***) | | |

**Table 5: Objective statistical results of Changhai Hospital with GT. Corrupted represents the original image before reconstruction, and proposed is the image processed using PERCEPT-Net. ns: not significant, *: $p < 0.05$, **: $p < 0.01$, ***: $p < 0.001$. All metrics were computed against ground truth. All comparisons are between the proposed method and corrupted input.**

| Type | Method | SSIM | PSNR | CNR | SNR | LPIPS | FID |
|---|---|---|---|---|---|---|---|
| T1 mild artifact | proposed | 0.717±0.132 (**) | 21.863±7.092 (***) | 55.148±0.252 (***) | 59.804±12.437 (***) | 0.199±0.085 (*) | 5.667±2.327 (**) |
| | corrupted | 0.670±0.092 | 19.328±3.085 | 35.457±0.114 | 36.345±1.516 | 0.221±0.068 | 5.969±2.058 |
| T1 moderate artifact | proposed | 0.754±0.151 (***) | 23.665±8.511 (***) | 48.026±0.260 (*) | 50.976±8.916 (*) | 0.186±0.093 (*) | 5.853±3.169 (*) |
| | corrupted | 0.609±0.043 | 18.260±2.389 | 32.261±0.138 | 35.460±2.009 | 0.270±0.032 | 10.895±3.558 |
| T1 severe artifact | proposed | 0.753±0.158 (***) | 24.667±7.320 (**) | 32.733±0.060 (*) | 39.534±0.840 (*) | 0.177±0.106 (*) | 7.946±8.748 (*) |
| | corrupted | 0.598±0.133 | 20.327±2.951 | 26.710±0.029 | 34.083±0.438 | 0.279±0.100 | 18.012±11.559 |
| T2 mild artifact | proposed | 0.645±0.172 (**) | 20.747±5.514 (**) | 33.047±0.228 (*) | 36.336±0.427 (*) | 0.180±0.082 (**) | 5.851±2.358 (*) |
| | corrupted | 0.607±0.131 | 19.692±3.962 | 33.728±0.206 | 35.419±2.935 | 0.198±0.064 | 5.894±1.271 |
| T2 moderate artifact | proposed | 0.602±0.134 (***) | 18.759±2.639 (**) | 31.633±0.221 (**) | 33.035±2.276 (**) | 0.206±0.080 (***) | 7.991±3.664 (***) |
| | corrupted | 0.476±0.109 | 18.333±2.110 | 25.558±0.180 | 32.740±3.128 | 0.296±0.064 | 12.358±2.573 |
| T2 severe artifact | proposed | 0.555±0.126 (***) | 17.404±2.419 (***) | 33.309±0.187 (***) | 32.956±2.593 (**) | 0.235±0.070 (***) | 9.159±2.782 (***) |
| | corrupted | 0.393±0.068 | 16.878±1.775 | 26.186±0.170 | 33.032±2.759 | 0.342±0.039 | 16.405±3.493 |

Table 6. Likert scale ratings of anatomical structure visibility and global diagnostic confidence, **Note: Ratings are reported on a** 1–5 Likert scale**, where 1 = poor visibility/low confidence, 5 = excellent visibility/high confidence. Data are presented as** median with minimum and maximum values in parentheses. **Significance was assessed using the Wilcoxon signed-rank test. ns: not significant,** *: $p < 0.05$, **: $p < 0.01$, ***: $p < 0.001$. **Comparisons are made between the proposed method and corrupted images.**

| Region | A1 | A2 | A3 | A4 | A5 | A6 | A7 | Global |
|---|---|---|---|---|---|---|---|---|
| Corrupted | 2 (1-4) | 2 (1-4) | 3 (1-4) | 3 (1-4) | 3 (1-4) | 2 (1-4) | 2 (1-4) | 2 (1-4) |
| Proposed | 3 (2-4) *** | 3 (2-4) *** | 3 (2-4) *** | 3 (2-4) *** | 3 (2-4) *** | 3 (2-4) *** | 3 (2-4) *** | 3 (2-4) *** |
| GT | 4 (4-5) | 4 (4-5) | 4 (4-5) | 4 (4-5) | 4 (4-5) | 4 (4-5) | 4 (4-5) | 4 (4-5) |